\pdfoutput=1
\documentclass{article}
\usepackage{amsmath, fancyhdr, courier, multirow, natbib}
\usepackage{amsmath, amsfonts, amssymb}
\usepackage[caption=false]{subfig}
\usepackage{graphicx}
\usepackage{color}

\usepackage{graphicx}

\usepackage[margin=1in]{geometry}
\begin{document} 
\title{
\fontsize{14pt}{1.2pt}{
\selectfont
\textbf{
		Empirical Evaluation of Kernel PCA Approximation Methods in Classification Tasks
	   }\\
       }
      }
\author{
	\fontsize{11pt}{1.2pt}{
		\selectfont{Deena P. Francis\textsuperscript{1}, Kumudha Raimond\textsuperscript{2}}}
	\rule[30pt]{0pt}{0pt}
       }
\date{\small
Department of Computer Sciences Technology,\\ Karunya Institute of Technology and Sciences, India\\
\texttt{\textsuperscript{1}deena.francis@gmail.com, \textsuperscript{2}kraimond@karunya.edu}}  	                                                                                                     	
\maketitle

\pagestyle{fancy}
\begin{changemargin}{0.5cm}{0.5cm} 
{\fontsize{9pt}{1.2pt}{
	\selectfont
\hspace{-0.4mm}\textbf{Abstract}. Kernel Principal Component Analysis (KPCA) is a popular dimensionality reduction technique with a wide range of applications. However, it suffers from the problem of poor scalability. Various approximation methods have been proposed in the past to overcome this problem. The Nystr\"{o}m method, Randomized Nonlinear Component Analysis (RNCA) and Streaming Kernel Principal Component Analysis (SKPCA) were proposed to deal with the scalability issue of KPCA. Despite having theoretical guarantees, their performance in real world learning tasks have not been explored previously. In this work the evaluation of SKPCA, RNCA and Nystr\"{o}m method for the task of classification is done for several real world datasets. The results obtained indicate that SKPCA based features gave much better classification accuracy when compared to the other methods for a very large dataset. Randomized Non-linear Component Analysis. Nystr\"{o}m method. Classification. Approximation}}
\end{changemargin}

\pagestyle{plain}
\section{Introduction}
\label{sec:intro}
Large input matrices are encountered in almost all modern applications. The large size is attributed to the large number of examples as well as large number of attributes. Performing any kind of analysis on them first requires some form of size reduction. Dimensionality reduction is technique employed to reduce the number of attributes in the data. It helps reduce the computational cost of applying traditional machine learning algorithms to such large data. Principal Component Analysis (PCA) \citep{pearson1901,hotelling1933analysis} is a very popular dimensionality reduction technique. Despite its popularity, it has a drawback. It assumes that the data are linearly separable. This assumption is not true in the case of all types of data. KPCA was proposed by \citet{scholkopf1997kernel} in order to overcome the problems of PCA. This method implicitly maps the data on to a higher dimensional space where the data becomes linearly separable. For the purpose of mapping, a kernel matrix $\mathbf{K}$ is constructed from the data. It contains the inner products between the input points in a higher dimensional feature space, $\mathbf{F}$. The explicit mapping onto any higher dimensional space is cleverly avoided by using the kernel trick. This trick makes use of certain 'kernel functions' which can be used to obtain the inner products between the data points in the new higher dimensional space. For an input matrix $\mathbf{A} \in \mathbb{R}^{n \times d}$, the kernel matrix is $\mathbf{K} \in \mathbb{R}^{n \times n}$, which stores the inner products between the points. KPCA has been widely used for applications such as face recognition \citep{wang2012kernel,kim2002face}, feature extraction and de-noising in non-linear regression \citep{rosipal2001kernel}, process monitoring \citep{lee2004nonlinear}, novelty detection \citep{hoffmann2007kernel}. 

KPCA suffers from the major drawback of poor scalability \citep{chin2007incremental}. When the number of input points increases, the space required to store $\mathbf{K}$ grows as $O(n^2)$, and the computational time of the algorithm grows as $O(n^3)$. So, it is no longer possible to store the whole kernel matrix in memory, also the time taken to perform KPCA on large data sets will be too high and thus becomes infeasible. In order to overcome the problem of huge computational cost of KPCA, many techniques have been proposed in the past. Nystr\"{o}m method approximated the eigen decomposition of a kernel matrix by sampling $m < n$ points from the input matrix and them computing the eigen decomposition on the smaller kernel matrix. The computational complexity of this method is $O(m^2n)$. RNCA is a combination of Random Fourier Features \citep{rahimi2008random} and PCA. The time and space taken by the algorithm is $O(m^n)$ and $O(dm)$ respectively. The space as well as computational complexity of KPCA can be reduced by using SKPCA \citep{ghashami2016streaming}. This algorithm maintained the principal components in a streaming fashion using a modified Frequent Directions (FD) algorithm \citep{liberty2013simple} and RFF. Its computational complexity is $O(ndm + nlm)$, where $l$ is the sketch size. The space required by the algorithm is $O(dm + lm)$.
\subsection{Main Contributions}
\label{sec:contributions}
The main contributions of this work are as follows.
\begin{itemize}
	\item Compare three KPCA approximation schemes in classification tasks.
	\item Study of the effect of algorithm specific parameters on the performance of the classifier.
	\item Provide classification results of applying the features generated by the algorithms on real world data sets.
	\item Analyze the results obtained, and provide their interpretation.
\end{itemize}
\subsection{Paper Organization}
\label{sec:org}
This work is organized as follows. The preliminary notations and formula are provided in Sect. \ref{sec:prelim}. The scalability issue of KPCA, approximate KPCA algorithms are described in Sect. \ref{sec:prelim}. In Sect. \ref{sec:method}, the methodology followed is described. The experimental results are provided in Sect. \ref{sec:exp} and discussion is provided in Sect. \ref{sec:dis}, and the conclusion is provided in Sect. \ref{sec:con}.
\section{Preliminaries}
\label{sec:prelim}
\subsection{Notations}
\label{sec:notations}
The input matrix $\mathbf{A}$ has size $n \times d$. A kernel matrix, $\mathbf{K}$ has size $n \times n$, and the kernel function is denoted by $\chi$. For any matrix $\mathbf{A} \in \mathbb{R}^{n \times d}$, $\mathbf{A}(i:j,j:k)$ denotes the subset of the matrix $\mathbf{A}$ having rows $i$ to $j$ and columns $j$ to $k$. $E[\mathbf{x}]$ denotes the expectation of a random variable $\mathbf{x}$. $\|\mathbf{A}\|_2$ denotes the spectral norm of a matrix $\mathbf{A}$, and is defined as the square root of the maximum eigenvalue of $\mathbf{A}^H \mathbf{A}$, where $\mathbf{A}^H$ is the conjugate transpose of $\mathbf{A}$. $\|\mathbf{A}\|_F = \sum_{i=1}^{n}|\mathbf{a}_i|^2$ is the Frobenius norm of $\mathbf{A}$. The pseudoinverse of $\mathbf{A}$ is denoted by $\mathbf{A}^{\dagger}$.
\subsection{KPCA}
\label{sec:kpca}
KPCA \citep{scholkopf1997kernel} is a non-linear version of PCA that can handle non-linearities in the data. The data is made to be linearly separable by mapping it to a higher dimensional space. A kernel matrix that contains the inner products between all the data points in the new higher dimensional space is constructed. For an input matrix $\mathbf{A}$ of size $n \times d$, the kernel matrix occupies size $n \times n$. Then the principal directions (vectors) that capture the maximum variance of the kernel matrix are computed. This is done by means of computing the eigen vectors of the kernel matrix. After the principal components are obtained, the newly arriving data are projected onto the subspace spanned by this new set of principal vectors. The computational complexity of the algorithm is $O(n^3)$ and the storage requirement of the algorithm is $O(n^2)$.
\subsection{Issues of KPCA}
\label{sec:issues}
KPCA requires computing and storing the kernel matrix $\mathbf{K}$ of size $n \times n$. In order to compute the kernel PCs, one one could resort to eigenvalue decomposition, which involves taking the Singular Vale Decomposition (SVD) of the kernel matrix, which takes time $O(n^3)$. The value of $n$ could be arbitrarily large and thus the time taken would be quite large. Thus it becomes infeasible to find the kernel PCs when the data is large. 
\section{Kernel PCA approximation methods}
\label{sec:approx}
In order to overcome the scalability issue of KPCA, many schemes have been proposed in the past. In this work we consider three most prominent of those. These three techniques have provable theoretical error guarantees, but their empirical performance in any real-world learning task has not been explored previously.
\subsection{Nystr\"{o}m KPCA}
\label{sec:nystrom}
The Nystr\"{o}m method for approximating the eigen decomposition of the kernel matrix was proposed by \citet{williams2001using}. This method involves uniformly sampling $m$ columns from the kernel matrix $\mathbf{K}$ to construct a low-rank kernel matrix $\tilde{\mathbf{K}}$. Note that sampling $m$ columns from $\mathbf{K}$ does not mean that we construct the full kernel matrix first and then sample. On the contrary, $m$ points are first sampled from the input matrix $\mathbf{X}$ and then the kernel function $\chi$ is applied on them.
\begin{equation}
\tilde{\mathbf{K}} = \mathbf{K}_{1:n,1:m}\mathbf{K}_{1:m,1:m}^{\dagger}\mathbf{K}_{1:m,1:n}
\label{eqn:nystrom1}
\end{equation}
where $m << n$.
The computational complexity of the Nystr\"{o}m method is $O(m^2 n)$. Although the original work by \citet{williams2001using} did not provide theoretical guarantees, a later work by \citet{wang2013improving} showed that the following bounds exist for the Nystr\"{o}m method. Here $\mathbf{K}_k$ is the best rank-$k$ approximation to $\mathbf{K}$.
\begin{equation}
\|\mathbf{K} - \tilde{\mathbf{K}}\|_F^2 \leq \left( \sqrt{1 + \frac{n^2k - m^3}{m^2(n-k)}} \right) \|\mathbf{K} - \mathbf{K}_k\|_F^2
\end{equation}
\begin{equation}
\|\mathbf{K} - \tilde{\mathbf{K}}\|_2^2 \leq \left( \frac{n}{m} \right) \|\mathbf{K} - \mathbf{K}_k\|_2^2
\end{equation}
\subsection{Randomized Nonlinear Component Analysis (RNCA)}
\label{sec:rnca}
RNCA \citep{lopez2014randomized} used Random Fourier Features (RFF) which was proposed by \citet{rahimi2008random}, followed by PCA. In RFF, the implicit mapping scheme in KPCA is replaced with an explicit mapping to a low-dimensional space. The implicit mapping function $\psi:\mathbb{R}^{d} \rightarrow \mathbb{R}^{m}$, where $m < d$ and $E[\psi(\mathbf{x})^T\psi(\mathbf{y})] = \chi(\mathbf{x}, \mathbf{y})$. First, $m$ feature maps, $z_i$, $i = 1, ..., m$ are obtained.
\begin{align*}
z_i(\mathbf{x}) &=  \left[ \cos(\boldsymbol{\alpha}_i^T \mathbf{x}_1 + b_i), \cos(\boldsymbol{\alpha}_i^T \mathbf{x}_2 + b_i), ..., \cos(\boldsymbol{\alpha}_i^T \mathbf{x}_n + b_i)\right], \quad z_i \in \mathbb{R}^{n}\\
	\psi_i &= \sqrt{\frac{2}{m}}z_i(\mathbf{x}) \\
	\mathbf{\Psi} &= [\boldsymbol{\psi}_1, \boldsymbol{\psi}_2, ...., \boldsymbol{\psi}_m], \quad \boldsymbol{\Psi} \in \mathbb{R}^{n \times m}
\end{align*}
where $\boldsymbol{\alpha}_i \in \mathbb{R}^{d}$ and it is obtained by sampling uniformly at random from the Fourier transform of the kernel function, and $b_i$ is drawn from the uniform distribution $\textit{Unif}(0,2\pi)$. RNCA works by constructing an approximation of the kernel matrix via the implicit mapping function and performing PCA on the result. The kernel matrix is obtained in the following manner, although not actually computed in practice.
\begin{equation}
\tilde{\mathbf{K}} = \mathbf{\Psi}\mathbf{\Psi}^T
\end{equation}
The error guarantee provided is as follows.
\begin{equation}
\mathbb{E}\|\mathbf{K} - \tilde{\mathbf{K}}\| \leq \frac{2n\log(n)}{m} + \sqrt{\frac{3n^2\log{n}}{m}}
\end{equation}
The computational complexity of RNCA is $O(m^2 n)$. The space required by the algorithm is $O(dm)$.
\subsection{SKPCA}
\label{sec:skpca}
SKPCA was proposed by \citet{ghashami2016streaming}. In this method, first the data is explicitly mapped to a low-dimensional space via Random Fourier Feature (RFF) maps \citep{rahimi2008random}. The function $\psi: \mathbb{R}^{d} \rightarrow \mathbb{R}^{m}$ is used to map each column vector of the input data matrix into the $m$-dimensional space. The features obtained are then given to a modified FD algorithm, which then returns the principal vectors. The space required by the algorithm is $O(dm + lm)$, training complexity is $O(ndm + lm)$, and testing time is $O(dm + lm)$. The approximation error obeys the bounds of Eqn. \ref{eqn:skpca_bound1} and Eqn. \ref{eqn:skpca_bound2}. For a small value of $\epsilon$, the approximate kernel matrix, $\tilde{\mathbf{K}} \in \mathbb{R}^{n \times n}$,
\begin{equation}
\|\mathbf{K} - \tilde{\mathbf{K}}\|_2 \leq \epsilon n
\label{eqn:skpca_bound1}
\end{equation}
\begin{equation}
\|\mathbf{K} - \tilde{\mathbf{K}}\|_F \leq \|\mathbf{K} - \mathbf{K}_k\|_F + \epsilon \sqrt{k}n
\label{eqn:skpca_bound2}
\end{equation}
This algorithm works in the streaming setting and is much faster than traditional KPCA.
\section{Methodology}
\label{sec:method}
\begin{figure}
	\centering
	\caption{Methodology}
	\label{fig1}
		\includegraphics[scale=0.55,trim=12mm 150mm 12mm 1mm,clip]{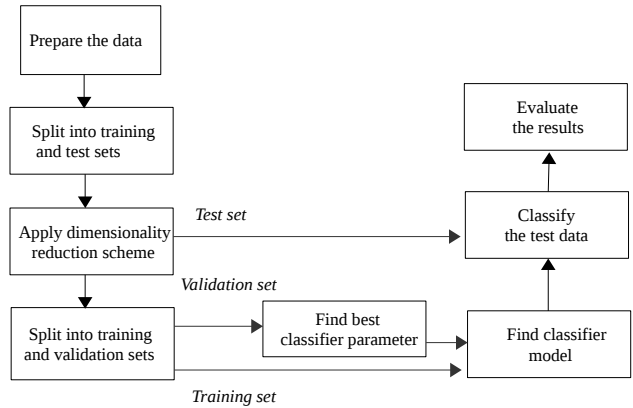}
\end{figure}

In this work the performance of the algorithms mentioned above are studied in the context of classification. The empirical performance of these algorithms for classification are tested using many datasets. The basic block diagram of the evaluation framework is shown in Fig. \ref{fig1}.
The data is first prepared for further processing and then split into training and test sets. Dimensionality reduction is applied on both sets. The best parameter for the classifier model is obtained after using a validation set which is a part of the training set. After the best parameter is found, the model for the classifier is obtained using the features extracted after applying either SKPCA, Nystr\"{o}m or RNCA method on the training data. Then the classifier model is used for predicting the class labels of the features after applying one of the above mentioned methods on the test data. Finally, the results are used for measuring the classification accuracy, precision, recall, and F-score \citep{sasaki2007truth}. F-score is a better measure when there is considerable skewness in the class distribution of the dataset. 
The classifier used in this work is Support Vector Machines (SVM) \citep{cortes1995support} which is a popular supervised classification algorithm. It has the soft margin parameter, $C$. In this work a linear SVM classifier is used, so the only parameter that needs to be tuned is C. This is done by means of the validation set and the training set as mentioned above. The toolbox of \citet{chang2011libsvm} was used for performing binary classification using SVM.
\section{Experiments}
\label{sec:exp}
All experiments were carried out in a Linux system with 4 GB RAM, Intel core i3 processor. The performance measures reported are only for the test dataset because the measures on the training set are not representative of the generalization performance of the classifier. The performance measures used are described below.
\begin{equation}
\text{Accuracy} = \frac{\text{Total no. of correctly classified instances}}{\text{Total no. of instances}}
\end{equation}
\begin{equation}
\text{Precision} = \frac{\text{TP}}{\text{TP+ FP}}
\end{equation}
\begin{equation}
\text{Recall} = \frac{\text{TP}}{\text{TP+ FN}}
\end{equation}
\begin{equation}
\text{F-score} = \frac{2 \times \text{Precision}\times \text{Recall}}{\text{Precision} + \text{Recall}}
\end{equation}
\begin{figure}
	\centering
	\subfloat []{\includegraphics[scale=0.4,trim=12mm 1mm 1mm 1mm,clip]{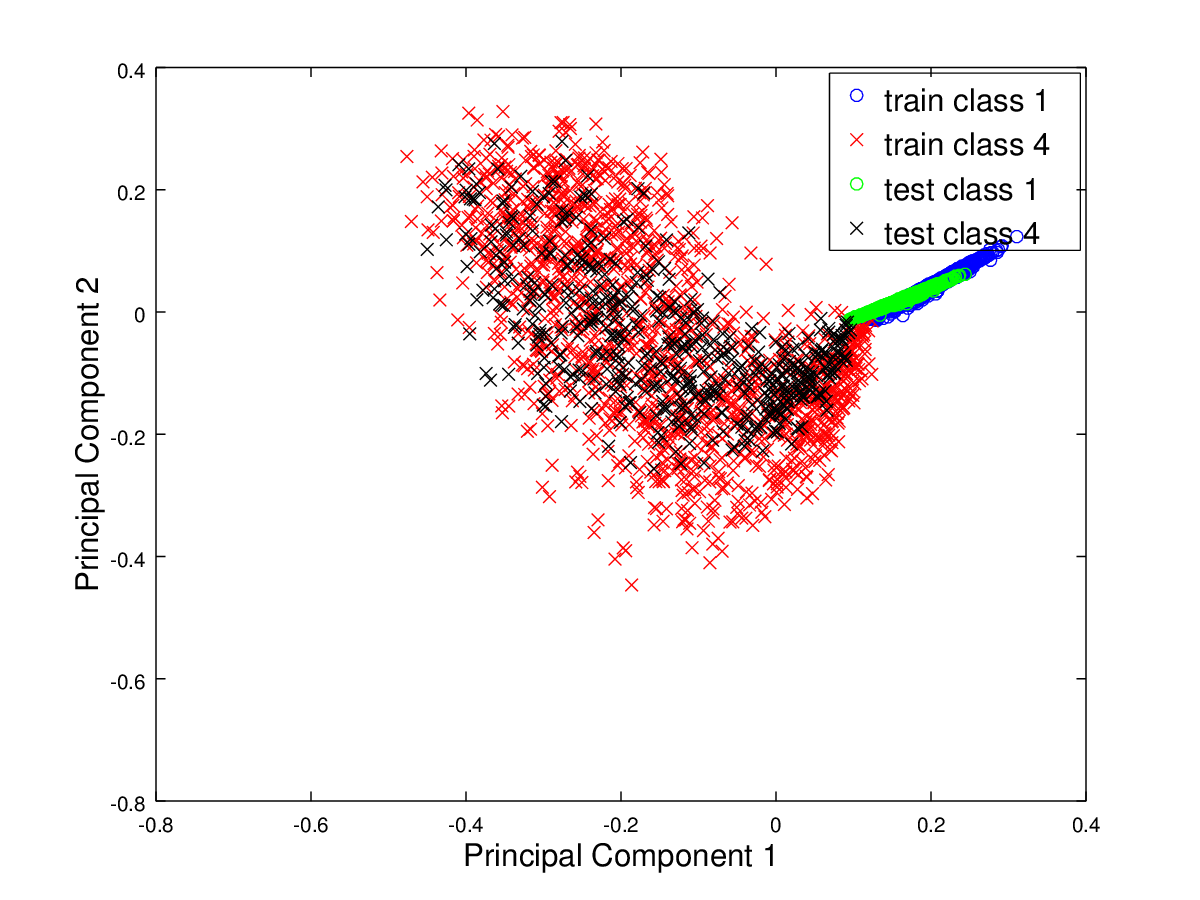}}
	\subfloat []{\includegraphics[scale=0.4,trim=10mm 1mm 1mm 1mm,clip]{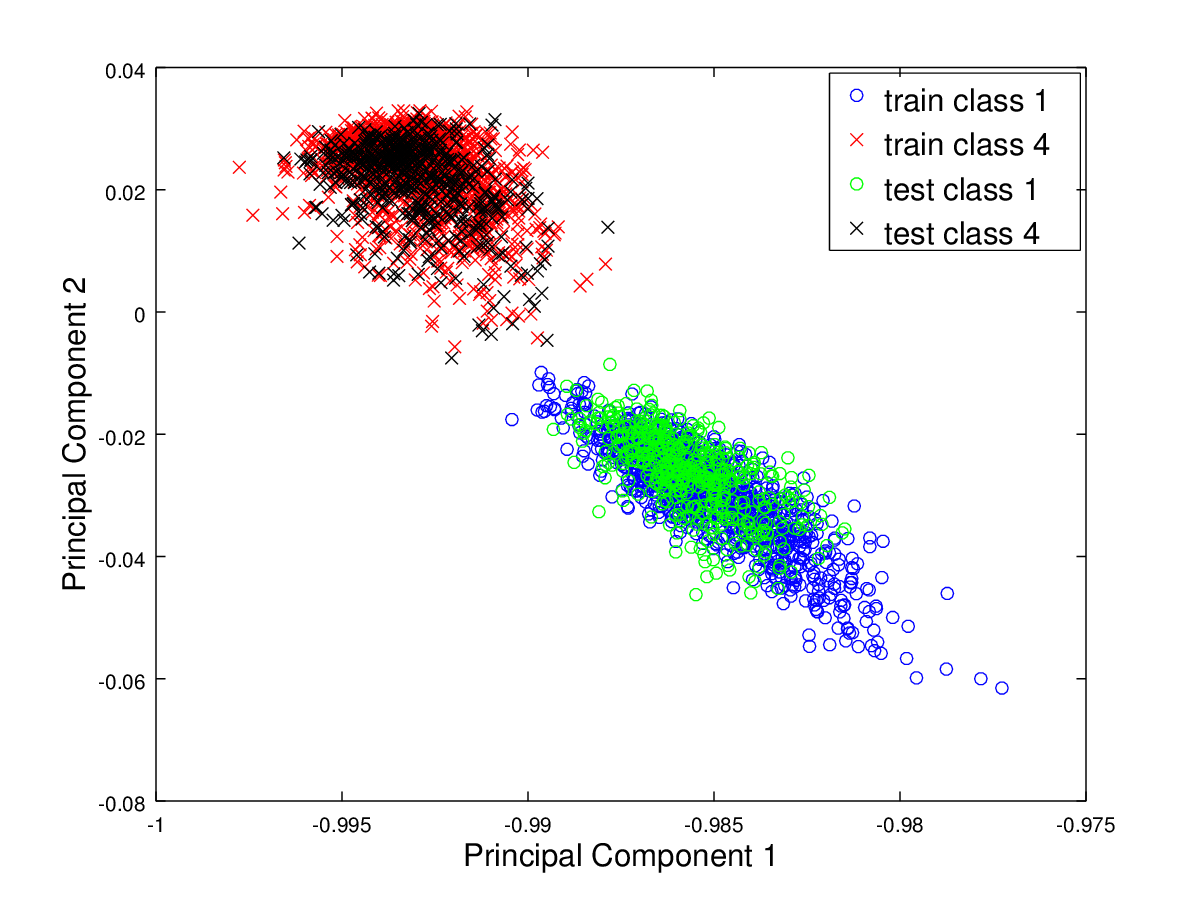}}\\
	\subfloat []{\includegraphics[scale=0.4,trim=6mm 1mm 1mm 1mm,clip]{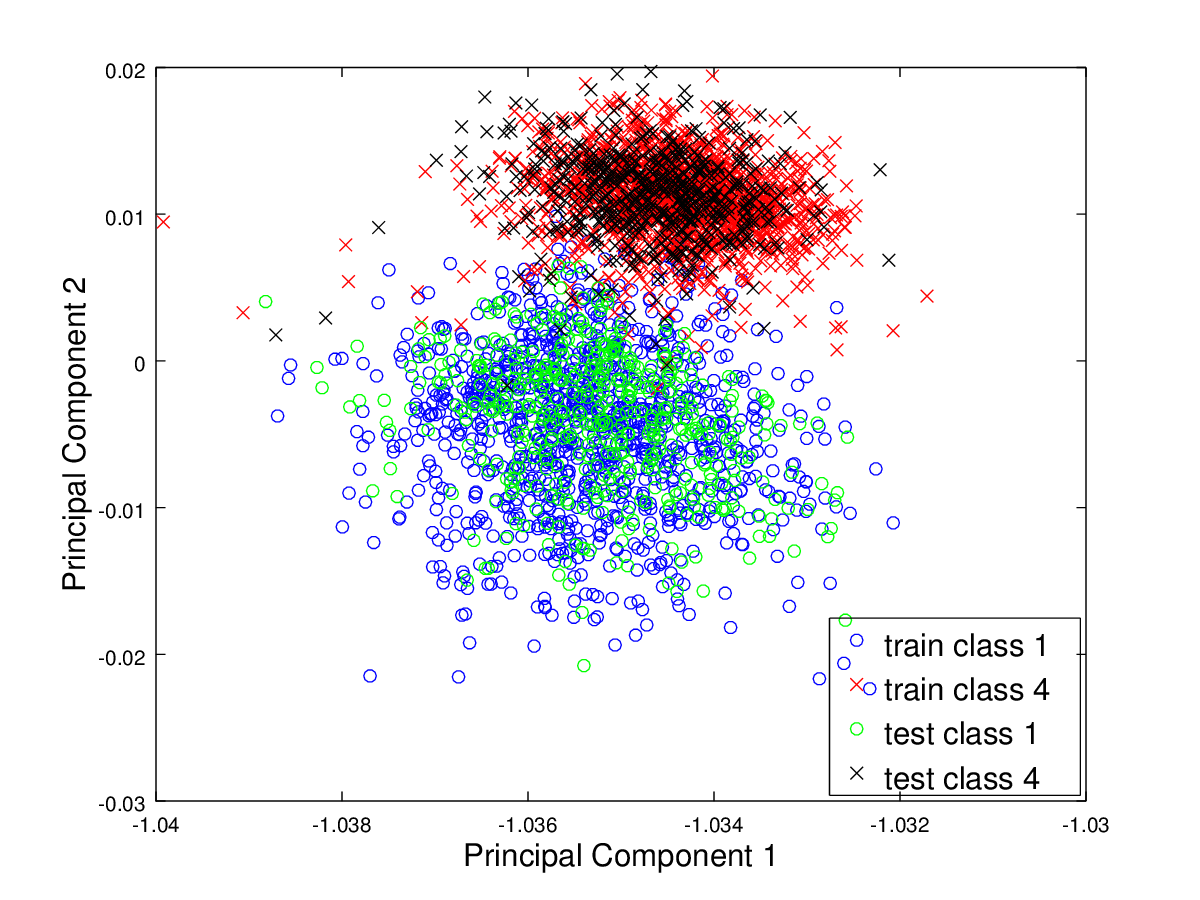}}
	\caption{Class separability of top two features obtained for HAR dataset via the three methods
		(a) Nystr\"{o}m-KPCA (b) RNCA (c) SKPCA}
	\label{fig2}
\end{figure}
\subsection{Datasets}
\label{sec:data}
\begin{itemize}
	\item MNIST \citep{lecun1998mnist}: contains images of handwritten digits $0,..,9$. The dataset has $10$ classes and the total number of instances is $60000$. Data belonging to classes $1$ and $7$ were used in this work and the number of instances of each class are $6742$ and $6265$ respectively. The size of the dataset used is $13007 \times 784$.
	\item HAR \citep{anguita2013public}: contains attributes pertaining to seven activities of 30 people. The label information is the type of activity each person is performing. The various activities (classes) of the dataset include walking, sitting, standing, laying, walking upstairs, walking downstairs, and standing. There are $7$ classes in the dataset with a total of $10299$ instances. Data belonging to classes $1$ and $4$ were used in this work. The number of instances of classes $1$ and $4$ are $1073$ and $1374$ respectively. The size of the dataset used is $2447 \times 561$.
	\item CIFAR \citep{krizhevsky2009learning}: contains a collection of color images from different categories like automobiles, trucks, birds, cats, dogs, deer, frog, horse, ship, and airplane. There are $10$ classes in this dataset with a total of $60000$. Data belonging to classes $7$ and $8$ were used in this work. The number of instances of classes $7$ and $8$ are $5000$ and $5000$ respectively. The size of the dataset is $10000 \times 3072$.
	\item ISOLET \citep{fanty1991spoken}: contains the recordings of the English alphabets spoken by $30$ speakers. There are $26$ classes in this dataset with a total of $7797$ instances. Data belonging to classes $1$ and $15$ were used in this work. The number of instances of classes $1$ and $15$ are $240$ and $240$ respectively. The size of the dataset used is $480 \times 617$.
\end{itemize}
\subsection{Feature Extraction}
\label{sec:feature_extraction}
Features are extracted from the data after applying SKPCA, Nystr\"{o}m-KPCA, or RNCA. The mapped points obtained after applying any of these methods should ideally be linearly separable in order to obtain good classification results. The visualization for the three methods are shown in Fig. \ref{fig2}. For ease of interpretation, the first two features (principal components) of all the mapped feature vectors are shown. Data used for the visualizations is the HAR dataset. It can be observed that for the HAR dataset, the separation is not clear and hence some error in classification can be expected.
\subsection{Spoken letter recognition}
\label{sec:spoken}
\begin{figure}[!h]
	\centering
	\includegraphics[scale=0.45,trim=6mm 1mm 1mm 1mm,clip]{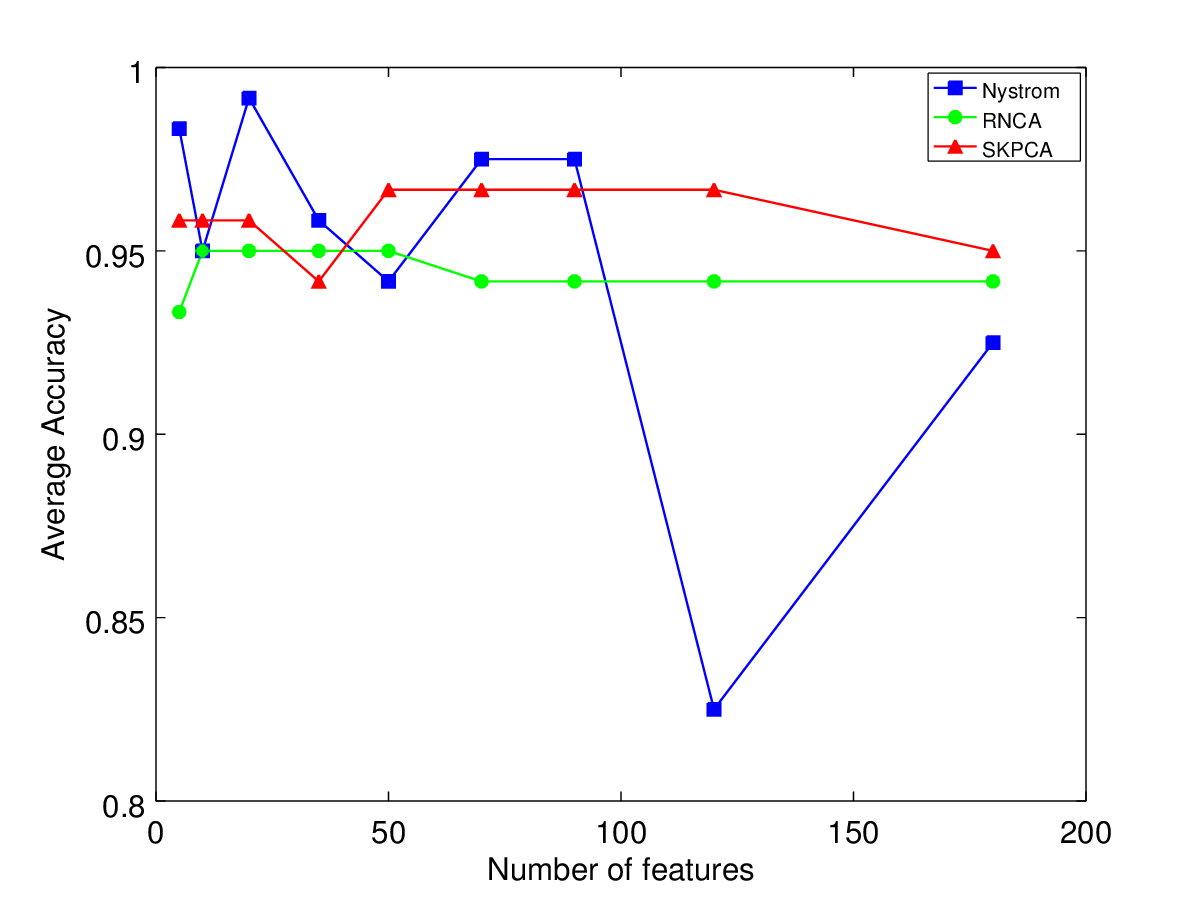}
	\caption{Variation of accuracy with number of features for ISOLET dataset}
	\label{fig3}
\end{figure}

For the task of spoken letter recognition, the ISOLET dataset from UCI repository was used. Here, binary classification was done, for which classes $1$ and $15$ (letters B and P) were used. The best accuracy obtained after using the features from the three methods for 50 features is shown in table \ref{tab1}. Fig. \ref{fig3} shows the variation of accuracy of SVM (linear kernel) with change in parameter (number of features) of the algorithms. When the number of features exceeds $100$, Nystr\"{o}m-KPCA based classification accuracy drops significantly. It can be observed that SKPCA does not perform better than Nystr\"{o}m- KPCA, but it performs better than RNCA for the ISOLET dataset.
\begin{table}[!h]
	\caption{Result of classification of ISOLET data using features obtained via the three methods}
	\label{tab1}
	\begin{center}
		\begin{tabular}{c c c c c}
			\hline
			\textbf{Method} & \textbf{Accuracy (\%)} & \textbf{Precision} & \textbf{Recall} & \textbf{F-score}\\
			\hline
			\hline
			Nystr\"{o}m KPCA & 94.16 & 0.94 & 0.94 & 0.94\\
			\hline
			RNCA & 95 & 0.95 & 0.95 & 0.95\\
			\hline
			SKPCA & 96.6 & 0.96 & 0.96 & 0.96\\
			\hline
		\end{tabular}
	\end{center}
\end{table}
\subsection{Human Activity Recognition (HAR)}
\label{sec:har}
\begin{figure}[!h]
	\centering
	\caption{Variation of accuracy with number of features for HAR dataset}
	\label{fig4}
	\includegraphics[scale=0.45,trim=1mm 1mm 1mm 1mm,clip]{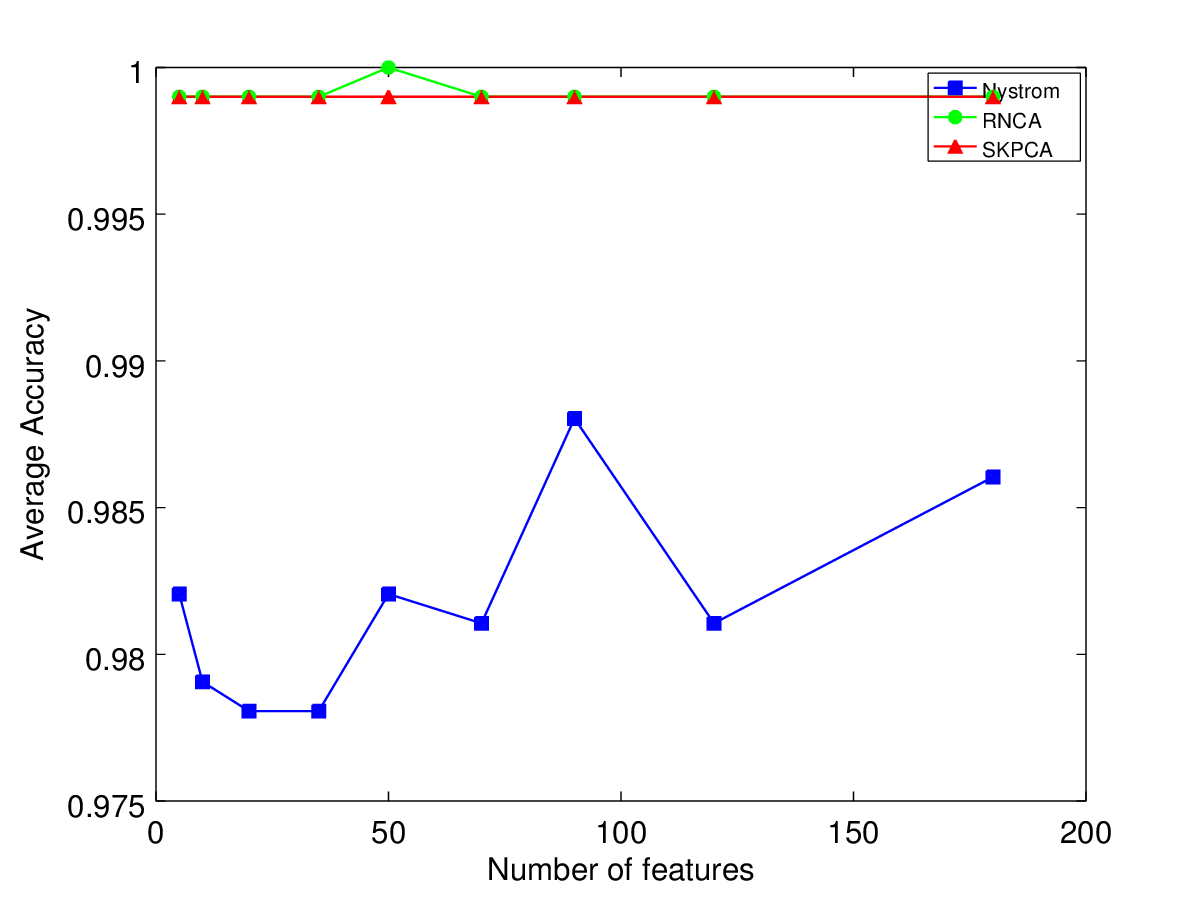}
\end{figure}

In this application, the movements of people recorded using a smartphone are used to classify them on the basis of different activities. For this application, the HAR dataset from the UCI repository was utilized. In the experiments, the classes 1 and 4 were utilized. Table \ref{tab2} shows the result of applying SVM classifier to the for 50 features generated by the three methods. Fig. \ref{fig4} shows the variation of accuracy of SVM (linear kernel) with change in parameter (number of features) of the algorithms. In the case of Nystr\"{o}m-KPCA, the accuracy keeps fluctuating, whereas in the case of SKPCA and RNCA, the accuracy remains almost constant for all parameter values, except for RNCA. This implies that only a small number of features is enough to
obtain good results for SKPCA as well as RNCA in the case of HAR dataset.
\begin{table}[!h]
	\caption{Result of classification of HAR data using features obtained via the three methods}
	\label{tab2}
	\begin{center}
		\begin{tabular}{c c c c c}
			\hline
			\textbf{Method} & \textbf{Accuracy (\%)} & \textbf{Precision} & \textbf{Recall} & \textbf{F-score}\\
			\hline
			\hline
			Nystr\"{o}m KPCA & 98.21 & 0.98 & 0.98 & 0.98\\
			\hline
			RNCA & 99.99 & 0.99 & 0.99 & 0.99\\
			\hline
			SKPCA & 99.99 & 0.99 & 0.99 & 0.99\\
			\hline
		\end{tabular}
	\end{center}
\end{table}
\subsection{Character Recognition}
\label{sec:char}
\begin{figure}[!h]
	\centering
	\includegraphics[scale=0.45,trim=1mm 1mm 1mm 1mm,clip]{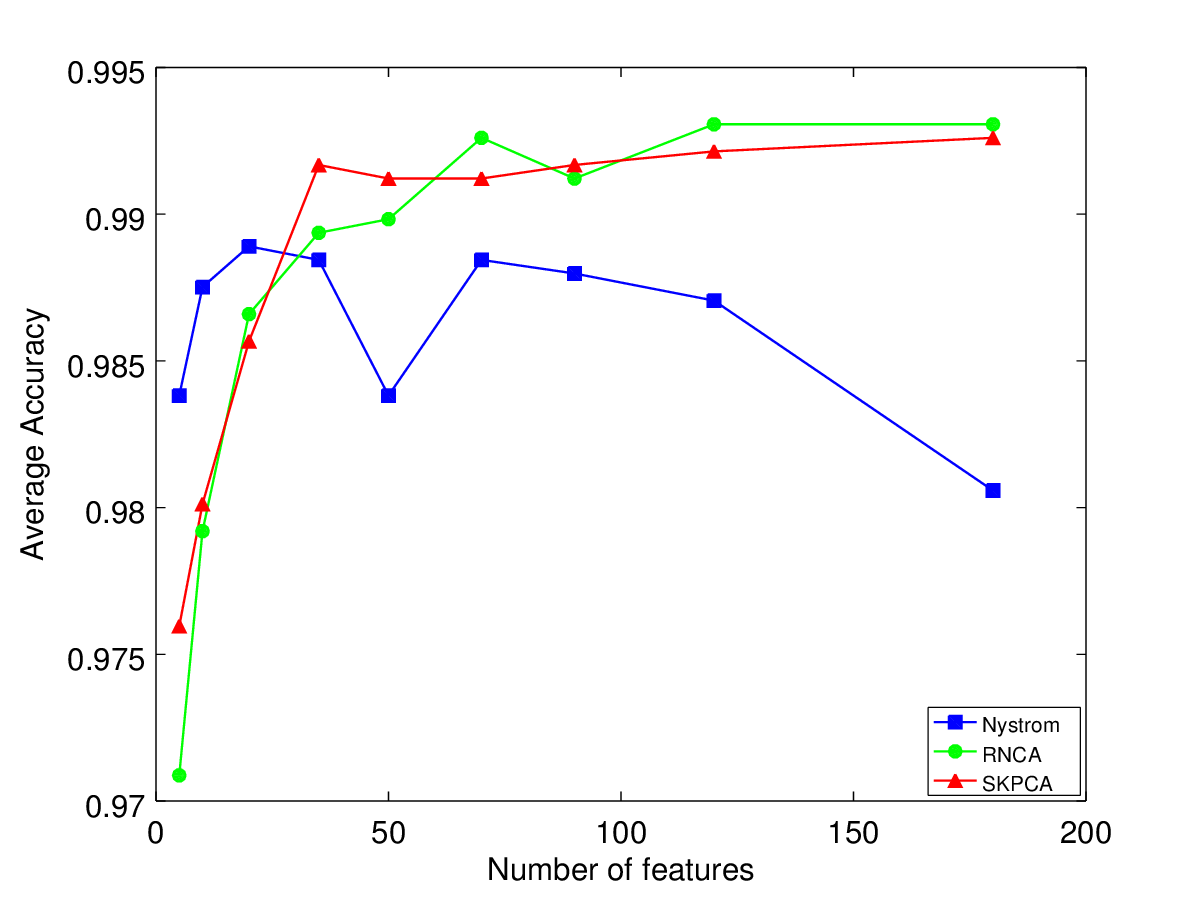}
	\caption{Variation of accuracy with number of features for MNIST dataset}
	\label{fig5}
\end{figure}

The task of character recognition requires distinguishing between two or more classes. This is generally a classification task. A widely used dataset used for this purpose is the MNIST dataset. For the task of binary classification of the MNIST data using classes 1 and 7, SVM classifier was used. The best values of accuracies obtained after using the three methods for 50 features is shown in table \ref{tab3}. Fig. \ref{fig5} shows the variation of accuracy of SVM (linear kernel) with change in parameter (number of features) of the algorithms.
\begin{table}
	\begin{center}
		\caption{Result of classification of MNIST data using features obtained via the three methods}
		\label{tab3}
		\begin{tabular}{c c c c c}
			\hline
			\textbf{Method} & \textbf{Accuracy (\%)} & \textbf{Precision} & \textbf{Recall} & \textbf{F-score}\\
			\hline
			\hline
			Nystr\"{o}m KPCA & 98.38 & 0.98 & 0.98 & 0.98\\
			\hline
			RNCA & 98.98 & 0.99 & 0.99 & 0.99\\
			\hline
			SKPCA & 99.12 & 0.99 & 0.99 & 0.99\\
			\hline
		\end{tabular}
	\end{center}
\end{table}

Although Nystr\"{o}m-KPCA based features obtain better accuracy for smaller number of features, the accuracy keeps decreasing after 60 features. In the case of SKPCA and RNCA, the accuracy increases as the number of features increases. In particular both the algorithms obtain better accuracy than Nystr\"{o}m-KPCA for more number of features. Thus, SKPCA and RNCA perform similarly in the case of MNIST dataset.
\subsection{Object classification}
\label{sec:obj}
\begin{figure}
	\centering
	\includegraphics[scale=0.45,trim=6mm 1mm 1mm 1mm,clip]{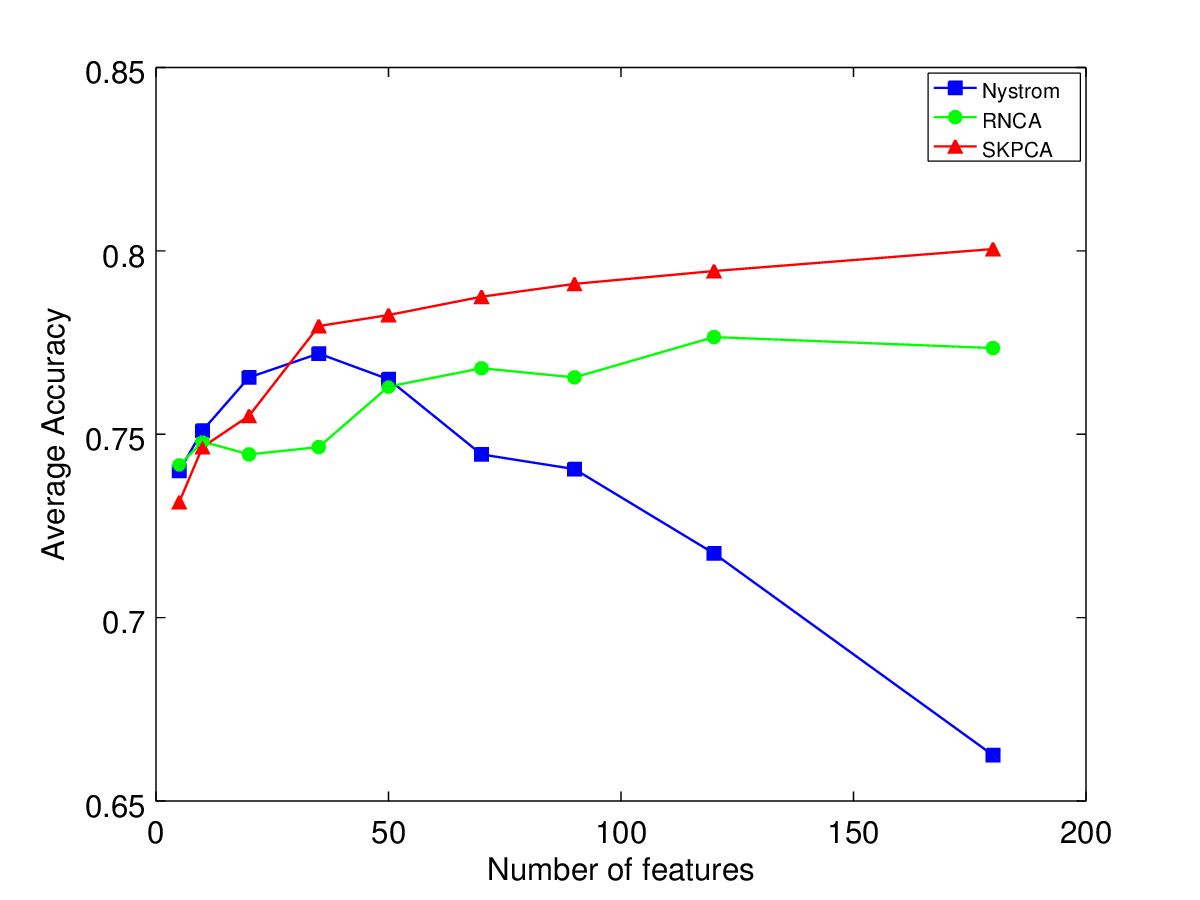}
	\caption{Variation of accuracy with number of features for CIFAR dataset}
	\label{fig6}
\end{figure}

The task of object classification requires predicting the class of a given object. For this application, the CIFAR dataset was used. In the experiments, the classes $7$ and $8$ (frog and horse) were used. The result of applying the for 50 features obtained after applying the three methods for the binary classification is shown in table \ref{tab4}. The variation of accuracy with the various parameter values (number of features) is shown in Fig. \ref{fig6}. In this dataset, SKPCA performs much better than RNCA and Nystr\"{o}m-KPCA on different parameter values. In fact the best accuracy of $80.05\%$ was obtained when SKPCA based features were used. One notable feature of this dataset is its large dimensionality $(d = 3072)$ and large number of samples $(n = 10000)$. Thus SKPCA is useful when the dataset is large.
\begin{table}
	\begin{center}
		\caption{Result of classification of CIFAR data using features obtained via the three methods}
		\label{tab4}
		\begin{tabular}{c c c c c}
			\hline
			\textbf{Method} & \textbf{Accuracy (\%)} & \textbf{Precision} & \textbf{Recall} & \textbf{F-score}\\
			\hline
			\hline
			Nystr\"{o}m KPCA & 76.50 & 0.76 & 0.76 & 0.76\\
			\hline
			RNCA & 76.30 & 0.76 & 0.76 & 0.76\\
			\hline
			SKPCA & 78.25 & 78.25 & 78.25 & 78.25\\
			\hline
		\end{tabular}
	\end{center}
\end{table}
\section{Discussion}
\label{sec:dis}
The results for binary classification of the MNIST dataset indicates that the SKPCA algorithm performs almost equally well as Nystr\"{o}m-KPCA and RNCA. SKPCA based features when used for classification provided results that are much better than the other two methods for a large dataset. The time required to generate the features for datasets are shown in table 5. It is clear that SKPCA has much lesser running time than Nystr\"{o}m-KPCA. For example, the time taken to find the features using Nystr\"{o}m method for the CIFAR data set is $88.48$s, whereas using SKPCA it is only $7.09$s. The running time of RNCA appears to be better than SKPCA, but the difference in running time becomes apparent only when the size of the dataset increases tremendously($n$ and $d$ of the order of millions). The running time of RNCA varies as $O(m^2 n)$, which is much too large when compared to $O(ndl + nlm)$ of SKPCA.
\begin{table}
	\caption{Time taken to generate the features}
	\label{tab:1}
	\begin{center}
		\begin{tabular}{c c c}
			\hline
			\rule{0pt}{2pt}\textbf{Dataset} & {\textbf{Algorithm}} & {\textbf{Time taken(s)}}\\
			\hline
			\hline
			\multirow{4}{*}{CIFAR}
			& {Nystr\"{o}m KPCA} & 88.48\\
			& {RNCA} & 1.27\\
			& {SKPCA} & 7.09\\
			\hline \rule{0pt}{1pt}
			\multirow{4}{*}{MNIST}
			& {Nystr\"{o}m KPCA} & 86.94\\
			& {RNCA} & 1.13\\
			& {SKPCA} & 6.39\\
			\hline \rule{0pt}{1pt}
			\multirow{4}{*}{ISOLET}
			& {Nystr\"{o}m KPCA} & 0.12\\
			& {RNCA} & 0.17\\
			& {SKPCA} & 0.32\\
			\hline \rule{0pt}{1pt}
			\multirow{4}{*}{HAR}
			& {Nystr\"{o}m KPCA} & 4.61\\
			& {RNCA} & 0.34\\
			& {SKPCA} & 1.24\\
			\hline
		\end{tabular}
	\end{center}
\end{table}
\section{Conclusion}
\label{sec:con}
In order to overcome the scalability issue of KPCA, various approximation schemes were proposed in the past. Nystr\"{o}m-KPCA, RNCA and SKPCA are three important KPCA approximation schemes. SKPCA overcomes the scalability issue of KPCA, and also provides a good approximation error bound. However the performance of these three algorithms in real world learning tasks has not been explored. In this work their performance in some real world applications such as character recognition, human activity prediction, spoken letter recognition, and object classification are studied. It was found that SKPCA features were indeed useful for classification and provides results very close to the other two methods, while requiring much lesser running time when compared to Nystr\"{o}m-KPCA. SKPCA based features when used for classification outperformed the other two methods for a very large dataset like CIFAR. Thus SKPCA can be used in learning tasks without much loss in accuracy for large datasets. Whereas in smaller datasets, applying RNCA or Nystr\"{o}m-KPCA would suffice.
\section*{Acknowledgements}
The authors would like to thank the financial support offered by the Ministry of Electronics and Information Technology (MeitY), Govt. of India under the Visvesvaraya PhD Scheme for Electronics and Information Technology.
\bibliographystyle{plainnat}
\bibliography{paper1}   
\end{document}